\documentclass[letterpaper, 10 pt, conference]{ieeeconf}  %

\IEEEoverridecommandlockouts                              %

\usepackage{layouts}

\usepackage{graphicx} %
\usepackage[caption=false, font=footnotesize]{subfig}
\usepackage{amsmath} %
\usepackage{amssymb}  %
\usepackage{bm}
\usepackage{cite}
\usepackage{algorithm}
\usepackage{algpseudocode}
\algnewcommand{\Initialize}[1]{%
  \State \textbf{Initialize:}
  \Statex \hspace*{\algorithmicindent}\parbox[t]{.8\linewidth}{\raggedright #1}
}

\usepackage{fancyhdr}

\title{\LARGE \bf
Robust Co-Design of Canonical Underactuated Systems for Increased Certifiable Stability
}

\author{Federico Girlanda$^{1,2}$, Lasse Shala$^{1}$, Shivesh Kumar$^{1,3}$ and Frank Kirchner$^{1,4}$%
\thanks{This work has been performed in the M-RoCK project funded by
the German Aerospace Center (DLR) with federal funds (Grant Number:
01IW21002) from the Federal Ministry of Education and Research (BMBF)
and is supported with project funds from the federal state of Bremen for
setting up the Underactuated Robotics Lab (Grant Number: 201-342-04-
2/2021-4-1).}%
\thanks{$^{1}$Robotics Innovation Center, German Research Center for Artificial Intelligence (DFKI GmbH), Bremen, Germany.}%
\thanks{$^{2}$Department of Information Engineering, University of Padova, Italy. {\tt\small federico.girlanda@outlook.com}}%
\thanks{$^{3}$Dynamics Division, Department of Mechanics \& Maritime Sciences, Chalmers University of Technology, Gotheburg, Sweden.}      
\thanks{$^{4}$Faculty of Mathematics and Computer Science, University of Bremen, Bremen, Germany.}%
}

\begin{document}

\renewcommand{\headrulewidth}{0pt}
\fancyhead[L]{\tiny © 2024 IEEE.  Personal use of this material is permitted.  Permission from IEEE must be obtained for all other uses, in any current or future media, including reprinting/republishing this material for advertising or promotional purposes, creating new collective works, for resale or redistribution to servers or lists, or reuse of any copyrighted component of this work in other works. \\ 2024 IEEE INTERNATIONAL CONFERENCE ON ROBOTICS AND AUTOMATION (ICRA). PREPRINT VERSION.}
\fancyfoot{}

\maketitle
\thispagestyle{fancy}
\pagestyle{empty}

\begin{abstract}
Optimal behaviours of a system to perform a specific task can be achieved by leveraging the coupling between trajectory optimization, stabilization, and design optimization. This approach is particularly advantageous for underactuated systems, which are systems that have fewer actuators than degrees of freedom and thus require for more elaborate control systems. This paper proposes a novel co-design algorithm, namely Robust Trajectory Control with Design optimization (RTC-D). An inner optimization layer (RTC) simultaneously performs direct transcription (DIRTRAN) to find a nominal trajectory while computing optimal hyperparameters for a stabilizing time-varying linear quadratic regulator (TVLQR). RTC-D augments RTC with a design optimization layer, maximizing the system's robustness through a time-varying Lyapunov-based region of attraction (ROA) analysis. This analysis provides a formal guarantee of stability for a set of off-nominal states. The proposed algorithm has been tested on two different underactuated systems: the torque-limited simple pendulum and the cart-pole. Extensive simulations of off-nominal initial conditions demonstrate improved robustness, while real-system experiments show increased insensitivity to torque disturbances.
\end{abstract}

\section{Introduction}
\label{sec_introduction}  
Different living organisms have evolved diverse locomotion strategies %
and physical characteristics to adapt to their unique environments. %
Similarly, for underactuated systems a combined
\begin{figure}[t]
   \centering
   \includegraphics[width=0.48\textwidth]{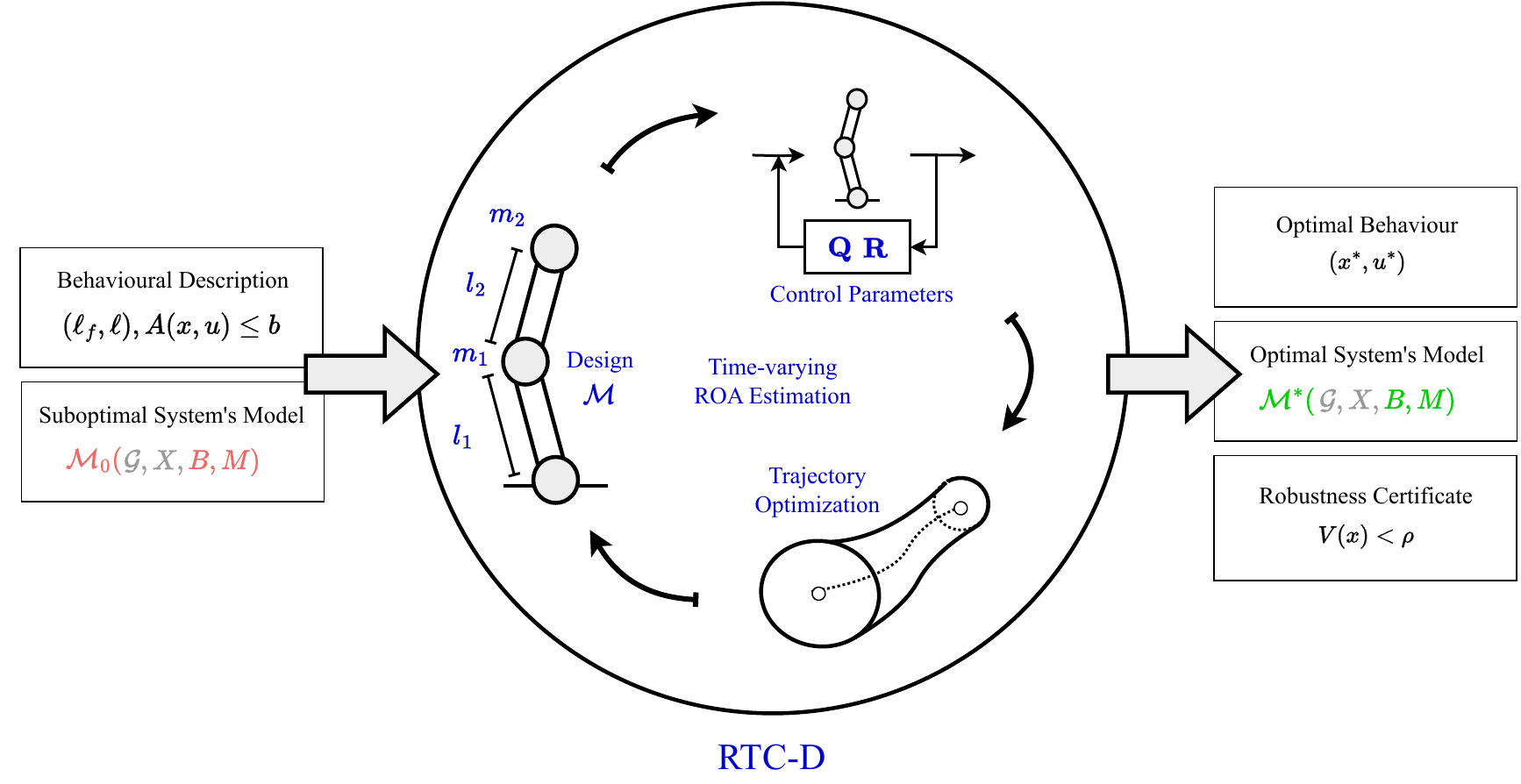}
   \caption{Robust co-optimization for the optimal fitness of the desired motion.}
   \label{fig:RobustCoDesign}
\end{figure}
 optimization of structural parameters and motion control is important to effectively accomplish the desired tasks. A design optimization tunes the system's hardware parameters, such as link lengths or the  center of mass position, of a system to allow for executing the described task with the desired behaviour. For trajectory tracking control, reaching the goal requires to search for a motion trajectory and to compute the control policy that permits the closed-loop trajectory following. Typically, the best trajectory is obtained via a trajectory optimization step, while the control input is computed by a specifically defined controller. A traditional approach to find the best trade-off between mechanical design and motion planning is to iterate between the two processes \cite{complexCoD}. Instead, concurrent design (co-design \cite{CoDintro}) aims to automate this process by numerically optimizing both the motion and design parameters, a strategy that has demonstrated superior results \cite{CoDex1, CoDex2, CoDex3, CoDex4, limbOpt, FadiniCoOpt, versatileCoD}. An example of \emph{gradient-based co-design} methods have been proposed in \cite{versatileCoD}. Here, the optimization for robot design and motion planning is handled on two levels. On the lower level, an efficient state-of-the-art constrained motion planner is implemented, which is continuously differentiable. This last property is then exploited on the higher level where the derivative of the motion is embedded into a nonlinear program. Another approach is the \emph{gradient-free co-design}. For instance, Ha et. al. \cite{limbOpt} proposed a framework that had successfully optimized designs for legged robots performing tasks that include jumping, walking, and climbing up a step. Similarly, Fadini et al. \cite{FadiniCoOpt} introduced a bi-level optimization scheme that finds the optimal actuator properties of a monoped in order to improve energy efficiency. Both approaches are two-staged and make use of variants of Monte-Carlo sampling to find candidate robot designs, which are subsequently evaluated through a motion planning stage. The covariance matrix adaption evolution strategy (CMA-ES) \cite{CMAES} is a commonly employed optimization strategy in this scenario. It uses a Gaussian prior on candidate design parameters and estimates a covariance matrix needed for the following sampling steps. Another popular gradient free algorithm is Nelder-Mead which has been used in the context of co-design of surgical robots~\cite{salunkhe2022efficient}.
 Recent studies in the literature are incorporating non-idealities, such as measurement noise and parameter variations, into co-optimization processes to reduce discrepancies between theoretical predictions and real-world performance \cite{robustMech, DIRTREL, FadiniRobustCoOpt, AcrobotCoDesign}. For instance, Manchester and Kuindersma \cite{DIRTREL} derived a tractable robust optimization algorithm, namely DIRTREL, that combines direct transcription with linear-quadratic control design to reason about closed-loop responses to disturbances. Fadini \cite{FadiniRobustCoOpt} introduced a simulation-based cost in a bi-level co-optimization algorithm that aims to take into account a given set of disturbances. The concept of robustness has also been explored through ROA analysis \cite{funnelIntro}, enabling the reasoning about stabilizable state sets. A formal mathematical certificate of stabilizability can also be provided. This methodology has mostly been applied for stability analysis but it's use in space-filling algorithms like LQR-trees has obtained promising results \cite{MooreLQRtrees, SOSLQRtrees, ReistSimROA}. ROA estimation has already proven valuable in co-design frameworks. In \cite{AcrobotCoDesign}, where the volume of the estimated region was used as an optimization cost for the up-right stabilization of an Acrobot system, resulting in a more robust closed-loop system against off-nominal initial states. See~\cite{2023_wiebe_doublependulum} for an extension of \cite{AcrobotCoDesign} to pendubot system. Additionaly, a bi-level optimization scheme was introduced by \cite{FadiniRobustCoOpt} to compute a robust energy efficient design and trajectory for a jumping monoped, considering a simulation-based robustness cost to enhance result's robustness.\\
This paper proposes a novel two layered \emph{gradient-free co-design} algorithm, namely RTC-D. The process involves co-optimization among the cost matrices of a TVLQR controller, the desired trajectory provided by DIRTRAN, and the system's design parameters. It aims to improve the real-world applicability of the combined optimization result by leveraging the volume of an estimated time-varying ROA. To the best knowledge of the authors this metric has not been studied in this context. We employ the CMA-ES optimization strategy to determine the optimal decision variables. Our approach has been rigorously tested on two different canonical underactuated systems: the torque-limited simple pendulum and a cart-pole. These systems are inherently underactuated, demanding a robust control approach. The specific case study involves the swing-up problem, consisting in stabilizing the pole in it's up-right position. We validate our results through extensive simulations of the system’s closed loop dynamics, assessing the stabilizability of off-nominal initial conditions. Furthermore, a real-world experiment has been implemented to test a scenario subject to input torque noise.
\paragraph*{Organization} Section~\ref{MathematicalBackground} introduces a mathematical description of the co-design problem and its components including trajectory optimization, TVLQR control and ROA estimation. Following this, Section~\ref{CodesignFramework} presents our proposed robust co-design methodology for canonical underactuated systems. Subsequently, in Section~\ref{ResultsAndDiscussion} we present results from applying this methodology to two different canonical systems namely torque limited pendulum and cart-pole system. Section \ref{ConclusionAndOutlook} concludes the paper. 
\section{Mathematical Background} 
\label{MathematicalBackground}
\subsection{Co-Design Problem Description}
\label{ProblemDescription}
A mechanical system with $n$ indepedent degrees of freedom can be defined by a system's description $\mathcal{M}(\mathcal{G}, \mathsf{X}, \mathsf{B}, \mathsf{M})$  where $\mathcal{G}$ is its topological graph, $\mathsf{X} = \text{diag}(\mathbf{X}_1, \ldots, \mathbf{X}_n) \in \mathbb{R}^{6n \times n}$ is the system matrix of body-fixed joint axis screw coordinates with $\mathbf{X}_i \in \mathbb{R}^{6}$, $\mathsf{B} = \{\mathbf{B}_1, \ldots, \mathbf{B}_n\}$ is the set of body frames with $\mathbf{B}_i \in SE(3)$ and $\mathsf{M} = \text{diag}(\mathbf{M}_1, \ldots, \mathbf{M}_n) \in \mathbb{R}^{6n \times 6n}$ is the  matrix of mass-inertia matrices of every moving body $(\mathbf{M}_i \in \mathbb{R}^{6 \times 6} )$ in $\mathcal{G}$. Given $\mathcal{M}(\mathcal{G}, \mathsf{X}, \mathsf{B}, \mathsf{M})$, it is straight forward to develop equations of motion (EOM) for any mechanical system in closed-form~\cite{2018_park_tutorial, mueller2021closed}. The EOM can be used to develop a 1st order ODE description of dynamics in terms of state-space $\mathbf{x} = (\mathbf{q}, \dot{\mathbf{q}}) \in \mathbb{R}^{2n}$ (with $\mathbf{q}$ and $\dot{\mathbf{q}}$ denoting the generalized position and velocity coordinates respectively) and action-space $\mathbf{u} \in \mathbb{R}^{p}$ in the form $\dot{\mathbf{x}} = f(\mathbf{x}, \mathbf{u})$. Given an initial state description, $\mathbf{x}_0$, and the mechanical system's model $\mathcal{M}$, a desired behavior can be achieved by optimizing its corresponding cost model, which is composed of a final cost model $l_f(\mathbf{x}_f)$ and running cost model $l\left(\mathbf{x}, \mathbf{u}\right)$, under a given set of constraints imposed by the robot and the environment. This typically entails solving an optimal control problem (OCP). In a typical OCP, the design parameters of $\mathcal{M}(\mathcal{G}, \mathsf{X}, \mathsf{B}, \mathsf{M})$ are assumed to be constant. 

We formulate co-design as a mathematical optimization problem with the decision variables including the minimal independent set of design space variables in $\mathcal{M}$ (while fixing the robot topology $\mathcal{G}$ and joint axes screws $\mathsf{X}$), state trajectory $\mathbf{x}(t)$, and control policy $\mathbf{u}(\mathbf{x}(t))$. The overall cost function of the co-design problem $(\ell_c, \ell_{cf})$ may include the behavior cost functions $l$ and $l_f$ as well as robustness of the control policy (e.g. volume of the region of attraction $\mathcal{B}$ of the controller). The problem can be mathematically written as the following.
\begin{equation}
    \label{trajOpt}
    \begin{aligned}
        &\underset{\mathcal{M}, \mathbf{x}(\cdot), \mathbf{u}(\cdot)}{\text{min}}&&\ell_{cf}(\mathcal{B}_f, l_f) + \int_{t_0}^{t_f} \ell_c(\mathcal{B}, l) dt\\
        &\text{subject to} &&\dot{\mathbf{x}}(t) = f(\mathcal{M}, \mathbf{x}(t), \mathbf{u}(t)),\quad \forall t \in \left[t_0, t_f\right]\\
        &&&\mathbf{x}(t_0) = \mathbf{x}_0, \mathbf{x}(t_0) = \mathbf{x}_N \\
        &&& \mathbf{x}_\text{min} \leq \mathbf{x} \leq \mathbf{x}_\text{max}, \mathbf{u}_\text{min} \leq \mathbf{u} \leq \mathbf{u}_\text{max} \\ 
        &&& \mathcal{M}_\text{min} \leq \mathcal{M} \leq \mathcal{M}_\text{max}
    \end{aligned}
\end{equation} 
As expected, this is a complex mathematical optimization problem with a very high dimensional space of decision variables. Hence, deliberation should go into decomposing the problem in a computationally tractable way. 
\subsection{Trajectory Optimization}
\label{sec_trajopt}
Trajectory optimization is a way of solving OCP where one solves for an open loop state and effort trajectory while optimizing the behavior cost model $(l_f, l)$. 
Direct methods discretize the trajectory optimization problem directly, converting it into a constrained parameter optimization problem \cite{tedrakeUnderactuated}. This discretization is also referred as \emph{transcription}. Transcription and the use of \begin{math}\mathbf{x}[\cdot], \mathbf{u}[\cdot]\end{math} as decision variables leads to the so-called direct transcription (DIRTRAN). The time discretization is done for $N$ points, the \emph{knot-points}.
\begin{equation*}
    \label{DIRTRAN}
    \begin{aligned}
        &\underset{\mathbf{x}\left[\cdot\right], \mathbf{u}\left[\cdot\right]}{\text{min}}&&l_f(\mathbf{x}\left[N\right]) + \sum_{k=0}^{N-1} l(\mathbf{x}\left[k\right], \mathbf{u}\left[k\right])\\
        &\text{subject to} &&\mathbf{x}\left[k+1\right] = f_d(\mathbf{x}\left[k\right], \mathbf{u}\left[k\right]),\quad \forall k \in \left[0, N-1\right]\\
        &&&\mathbf{x}\left[0\right] = \mathbf{x}_0\\
        &&& \mathbf{x}_\text{min} \leq \mathbf{x} \leq \mathbf{x}_\text{max}, \mathbf{u}_\text{min} \leq \mathbf{u} \leq \mathbf{u}_\text{max} 
    \end{aligned}
\end{equation*}
The resulting open-loop trajectory $(\mathbf{x}^{\star}, \mathbf{u}^{\star})$, usually referred as \emph{nominal trajectory}, has to satisfy the discrete dynamic's $f_d$ constraints while minimizing the standard additive-cost optimal control objective $l = (\mathbf{x}[k]-\mathbf{x}_N)^T \mathbf{Q}_{T} (\mathbf{x}[k]-\mathbf{x}_N) + \mathbf{u}^T \mathbf{R}_T \mathbf{u}$ and final cost $l_f = (\mathbf{x}[k]-\mathbf{x}_N)^T \mathbf{Q}_{Tf} (\mathbf{x}[k]-\mathbf{x}_N)$ where $(\mathbf{Q}_T, \mathbf{Q}_{Tf})$ denote the running and final state regularization cost matrices respectively and $\mathbf{R}_T$ denotes the effort minimization cost matrix to reach the goal state $\mathbf{x}_N \in \mathbb{R}^{2n}$. This optimization problem can be solved using commercial sequential-quadratic programming (SQP) solvers, such as SNOPT \cite{SNOPT}. The set of hyperparameters of DIRTRAN is defined as the set $\mathcal{H}_T = \{\mathbf{Q}_T, \mathbf{R}_T, \mathbf{Q}_{Tf}\}$.

\subsection{Trajectory Stabilization}
\label{sec_trajstab}
The stabilization of a nominal trajectory defined in a finite time interval \begin{math}t \in \left[t_0, t_f\right]\end{math} can be achieved via \emph{Time-Varying LQR (TVLQR)}~\cite{tedrakeUnderactuated}. 
TVLQR aims to minimize the error coordinates $\mathbf{\bar{x}} = (\mathbf{x} - \mathbf{x}^*)$ and $\mathbf{\bar{u}} = (\mathbf{u} - \mathbf{u}^*)$ 
, where ${}^*$ denote states of the nominal trajectory obtained from trajectory optimization step. 
For this, a time-varying linearization using a Taylor series approximation is performed, resulting in a time-varying linear system in the error coordinates:
\vspace{-0.1cm}
\begin{equation}
\label{eq:timevarylinearsys}
\mathbf{\dot{x}} = \mathbf{A}(t) \mathbf{\bar{x}}(t) - \mathbf{B}(t) \mathbf{\bar{u}}(t)
\vspace{-0.1cm}
\end{equation}
The quadratic cost function is defined as:
\vspace{-0.1cm}
\begin{equation*}
J = \mathbf{\bar{x}}^T(t) \mathbf{Q}_{Cf} \mathbf{\bar{x}}(t) + \int_{0}^{t_f} \left( \mathbf{\bar{x}}^T(t) \mathbf{Q}_C \mathbf{\bar{x}}(t) + \mathbf{\bar{u}}^T(t) \mathbf{R}_C \mathbf{\bar{u}}(t)\right) dt
\vspace{-0.1cm}
\end{equation*}
where $\mathbf{Q}_C=\mathbf{Q}_C^T \succeq 0$, $\mathbf{Q}_{Cf}=\mathbf{Q}_{Cf}^T \succeq 0$ and $\mathbf{R}_C=\mathbf{R}_C^T \succ 0$. 
The optimal cost-to-go can be written as a time-varying quadratic term and the controller gain $\mathbf{K}(t)$ be found by solving the differential Riccati Equation (DRE).
The optimal tracking control policy is given by:
\begin{equation*}
    \begin{aligned}
        &\pi(\mathbf{x}, \mathbf{u}) & = \quad& \mathbf{u}^{\star}(t) - \mathbf{K}(t) \bar{\mathbf{x}}\\
        & & = \quad& \mathbf{u}^{\star}(t) - \mathbf{R}_C^{-1}\mathbf{B}^T\mathbf{S}(t)(\mathbf{x} - \mathbf{x}^{\star}(t))
    \end{aligned}
\end{equation*}
The solution of the DRE provides the optimal cost-to-go $J^{\star}(\mathbf{x},t) = \bar{\mathbf{x}}^T \mathbf{S}(t) \bar{\mathbf{x}}$, a function that returns the accumulated cost when running the optimal controller from any initial state to the goal. The set of hyperparameters of TVLQR controller is defined as the set $\mathcal{H}_C = \{\mathbf{Q}_C, \mathbf{R}_C, \mathbf{Q}_{Cf}\}$.
\subsection{Region of Attraction Estimation}
\label{sec_roaest}
The region of attraction of a closed-loop system is defined as the set of states around a fixed point that are guaranteed to be stabilized by the controller. Most often, it is difficult, if not impossible, to determine analytically and an estimation has to be considered. The estimation methods which are available in the literature can be classified in two main categories: Lyapunov-based and Non-Lyapunov methods \cite{Najafi2016}.
Lyapunov analysis can be employed to obtain a formal guarantee of stability for an inner estimate of the ROA. The Lyapunov function $V$ is a generic function of the state such that 
\[V(\mathbf{x}^{\star}) = 0 \quad \text{and} \quad V(\mathbf{x}) > 0,\ \dot{V}(\mathbf{x}) < 0 \quad \forall \mathbf{x} \in \mathcal{X}\backslash \{\mathbf{x}^{\star}\}\]
If the above conditions hold for the fixed point $\mathbf{x}^{\star}$, then it is globally asymptotically stable. Sublevel sets $\mathcal{B} = \{\mathbf{x} | V(\mathbf{x}) < \rho\}$, $\rho > 0$, of a Lyapunov function are then used as approximations of the region of attraction \cite{KhalisBook}. For closed loop dynamics under LQR control, the cost-to-go matrix $\mathbf{S}$ can be used to construct a suitable quadratic Lyapunov function. Setting $V = \bar{\mathbf{x}}^T \mathbf{S} \bar{\mathbf{x}}$ makes the estimation to be a matter of finding a suitable $\rho$. In the time-varying case each knot point of the nominal trajectory is associated to a ROA, hence a vector $\boldsymbol{\rho} \in \mathcal{R}^N_{+}$ has to be estimated. In this case, the resulting ROA is also called \emph{funnel} \cite{funnelIntro}. Two methods for computing the funnel are described below:
\subsubsection{Sum of Squares (SOS)}
In the case of polynomial dynamics, SOS optimization can be used to express the estimation task as an optimization problem. Polynomial, non-linear dynamics can always be obtained via Taylor approximation. The idea is that every initial state state taken in the estimated ROA of the knot point $k$ must end up, following the system's dynamics, inside the ROA estimate associated to $k+1$. An estimation procedure for each couple of knot points is then considered. Each estimation process is a bilinear alternation between two optimizations~\cite{Moore_2014}:
\begin{equation*}
    \label{SOS_tv}
    \begin{aligned}
        &\text{Multiplier Step:} \\ 
        &\underset{\gamma_i, \lambda_i}{\text{max}}&&\gamma_i\\
        &\text{subject to} &&-(\dot{V}-\dot{\boldsymbol{\rho}}_i) + \lambda_i (V - \boldsymbol{\rho}_i) - \gamma_i\ \ \ is\ SOS\\
        &&&\lambda_i \quad is\ SOS\\
        &&&\gamma_i > 0 \\ 
        &\text{Rho Step:} \\ 
        &\underset{\boldsymbol{\rho}_i}{\text{max}}&&\boldsymbol{\rho}_i\\
        &\text{subject to} &&-(\dot{V}-\dot{\boldsymbol{\rho}}_i) + \lambda_i (V - \boldsymbol{\rho}_i) \ \ \ is\ SOS\\
        &&&\lambda_i \quad is\ SOS\\
        &&&\boldsymbol{\rho}_i > 0
    \end{aligned}
\end{equation*}
Given a meaningful initial guess of $\boldsymbol{\rho}$, these two steps have to be alternatively solved until a convergence condition. This condition can be, for instance, related to the overall magnitude of $\boldsymbol{\rho}$ . After a time-invariant $\boldsymbol{\rho}_N$ has been determined, a bilinear alternation for each couple of knot points has to be solved going backwards for $k \in [0, N-1]$. By means of the S-procedure \cite{Sproc}, we are constraining the Lyapunov function to change slower than $\boldsymbol{\rho}$ for all the states at the boundary of the ROA. This is sufficient for the dynamics to completely stay inside the funnel~\cite{tedrakeUnderactuated}.
\subsubsection{Simulation based}
The methodology proposed in \cite{ReistSimROA} is based on simulations and falsifications. Initially, a funnel hypothesis is proposed, then each trajectory knot point is verified through sampling and simulations. Given a desired trajectory $(\mathbf{x}^*,\mathbf{u}^*)$ defined for $N$ knot points, the initial ROA guess $k$ is set as an open ball with radius $\rho_k > 0$, centered at $\bar{\mathbf{x}}_k$:
\[\mathcal{B}(\bar{\mathbf{x}}_k , \rho_k ) := \{\mathbf{x}_k : V(\bar{\mathbf{x}}_n) < \rho_n\}.\] Assume that a stabilizable region $\mathcal{B}_N$ around the goal state is given. For each node a set of samples is obtained and simulated through the end of the nominal trajectory. A simulation is considered successful if the sampled initial condition has been stabilized to the goal region, $\mathbf{x}_N \in \mathcal{B}_N$. If so, the verified ROA remains valid. If the simulation fails the initial region guess and the following ones are shrunk accordingly to the following rule:
\[\rho_{i,new} = \text{min}(V(\bar{\mathbf{x}}_i ), \rho_{i,old} ),\ \forall i \in \left[k, N-1\right].\]
\section{Proposed Co-design Framework} 
\label{CodesignFramework}
In order to solve the problem described in Section \ref{ProblemDescription}, a bi-level co-optimization algorithm has been conceived which aims to seek shared optimality between design optimization, trajectory optimization and trajectory stabilization with the aim of optimizing the final robustness of the control scheme. The trajectory optimization and stabilization tasks have been solved through DIRTRAN and TVLQR control, respectively.  This allows the introduction of explicit constraints on the motion variables, which is not possible in Differential Dynamic Programming (DDP) where the constraints are introduced as weighted costs in the optimization.
\begin{figure}[ht]
    \centering
    \includegraphics[width=0.49\textwidth]{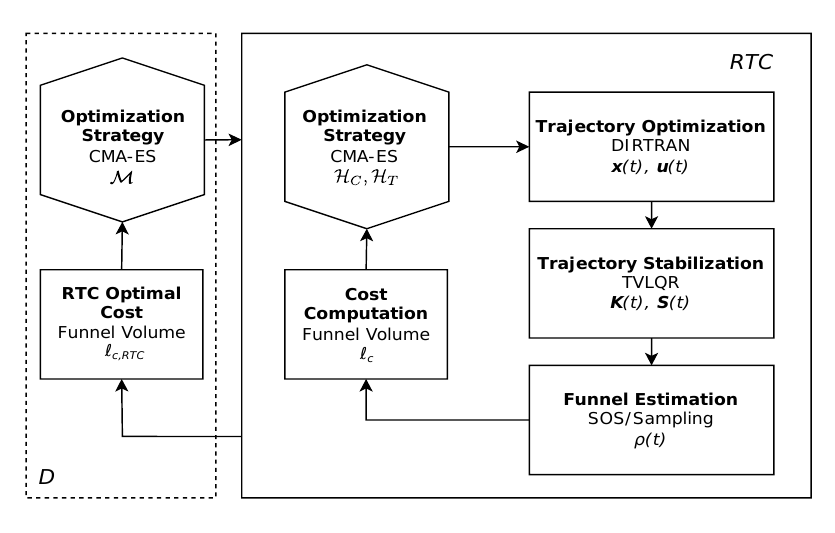}
    \caption{RTC-D algorithm scheme.}
    \label{fig:RTCDscheme}
\end{figure}\\
These two choices are also compatible with the funnel estimation process. The volume of the time-varying ROA was chosen as a metric for the system's robustness and calculated as the sum of the volumes of the ROAs at each knot point. The gradient-free CMA-ES method has been chosen to optimize the decision variables. This method operates by adapting the covariance matrix of the search distribution for a population of candidate solutions, which are referred to as individuals. CMA-ES does this efficiently with limited evaluations, giving preference to individuals that perform better during the optimization process.

\subsection{RTC}
The inner-layer of the proposed algorithm, namely $RTC$, is concurrently optimizing the nominal trajectory and the stabilizing controller, for fixed design parameters $\mathcal{M}$. This has been implemented by considering the hyperparameters of the trajectory optimization $\mathcal{H}_T$ and stabilizing controller $\mathcal{H}_C$ as decision variables for the optimization problem. 
We have assumed the decision variables to stay inside some reasonable bounds to restrict the search space.
At each iteration, the optimizer proposes a new set of hyperparameter cost matrices. They are used to compute a new nominal trajectory $(\mathbf{x}^*, \mathbf{u}^*)$ via DIRTRAN method as explained in Section~\ref{sec_trajopt} and a stabilizing control policy $\mathbf{\pi}^*$ via TVLQR method as explained in Section~\ref{sec_trajstab}. These quantities are necessary to compute the ROA of the stabilizing controller. Finally, the volume $\ell_c$ of the estimated funnel is used by the CMA-ES strategy to weigh each initially proposed set of costs and provide better $\mathcal{H}_T, \mathcal{H}_C$ such that the volume of the resulting ROA is maximized. We iterate this process until a predefined maximum number of evaluations of the objective function.
\subsection{RTC-D}
To complete the solution of our co-design problem, an outer CMA-ES optimization layer, namely $D$, with the capability to vary the design parameters in $\mathcal{M}$ has been added. Specifically, we consider the minimal independent set of design space variables in $\mathcal{M}$ (while fixing the robot topology $\mathcal{G}$ and joint axes screws $\mathsf{X}$). Also, in this case the decision variables are constrained inside some reasonable bounds which may be inspired from manufacturing constraints or availability of off-the-shelf parts. Compared to $RTC$, the resulting process is more computationally expensive due to the increased number of optimization layers. On the other hand, the solver has now more power to improve the objective function. 
The optimization strategy proposes a new set of design parameters for each iteration. The $RTC$ layer fixes them and then determines an optimal stabilized trajectory along with the related funnel volume. A different set of design parameters is then provided by the $D$ CMA-ES optimizer to perform the same operation. Eventually, designs that maximize the ROA volume computed in the inner layer will be preferred by the solver. Similar to RTC, we iterate this process until a predefined maximum number of objective function evaluations.

\section{Results and Discussion} 
\label{ResultsAndDiscussion}
\begin{figure}[ht]
    \centering
    \includegraphics[width=0.45\linewidth]{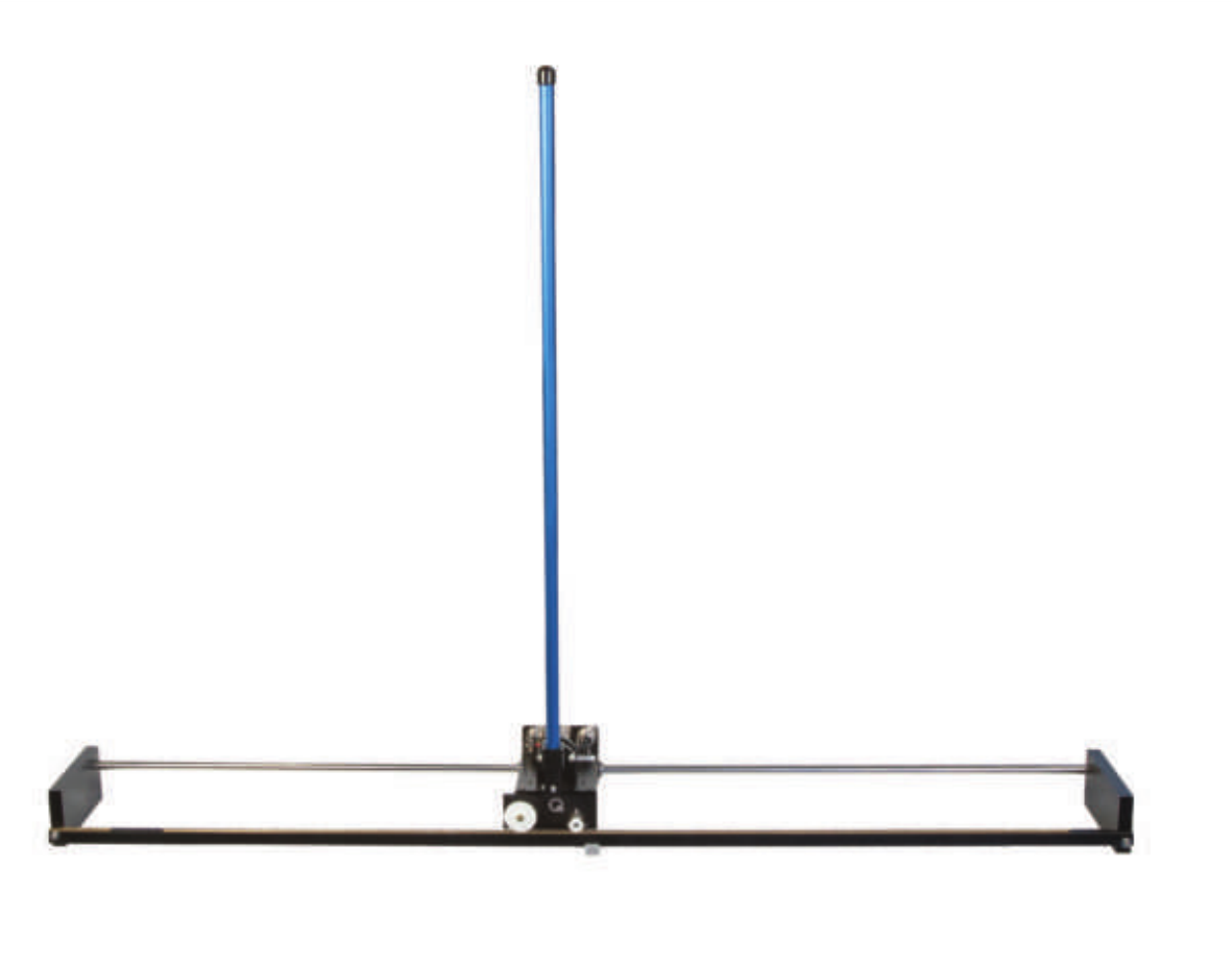}
    \includegraphics[width=0.45\linewidth]{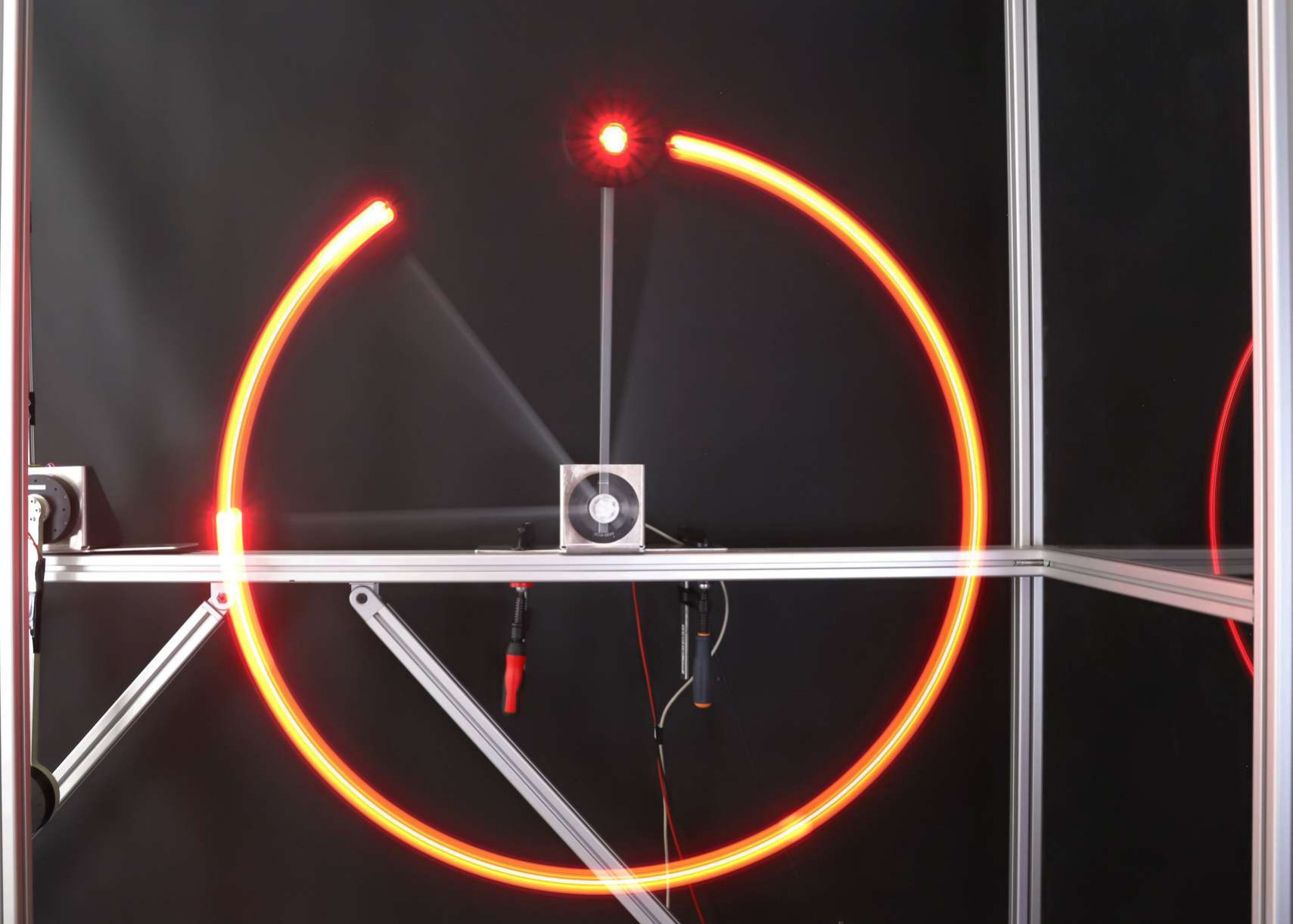}
    \caption{Experimental Systems: Cart Pole (left) and Simple Pendulum (right)}
    \label{fig:realSys}
\end{figure}

In this section the optimization results and their verification is presented\footnote{The open-source code implementation is available at https://github.com/dfki-ric-underactuated-lab/robust\_codesign}. Two underactuated systems have been considered for simulated and experimental verification: torque-limited simple pendulum~\cite{DFKIsp} and cart-pole~\cite{QuanserLIP} (Figure \ref{fig:realSys}). The common task is to solve the well-known swing-up problem. In particular, the optimization parameters were inspired by the real hardware available in the \emph{Underactuated Robotics Lab} of the DFKI Robotics Innovation Center. 
The computations related to the simple pendulum have been handled by a having 2x 8-core Xeon E5-2630 v3 (Haswell) machine with 128GB of ECC RAM. The computations for the cart-pole have been performed by a 2-core 4-thread 2.70 GHz Intel(R) Core(TM) i7-7500U computer with 8 Gb of RAM. A maximum of 3 parallelized cost computations have been considered for both the systems.

The controller and trajectory cost matrices have been assumed to be the same  i.e. $\mathbf{Q}_{T} = \mathbf{Q}_{C} = \mathbf{Q}  = \text{diag} (Q_{11}, \ldots, Q_{2n2n})$ and $\mathbf{R}_{T} = \mathbf{R}_{C} = \mathbf{R}  = \text{diag} (R_{11}, \ldots, R_{pp})$ during the optimization in order to decrease the overall number of decision variables and hence to reduce the computational complexity. For simplicity, the final cost matrices for the trajectory optimization and TVLQR controller were fixed as $\mathbf{Q}_{Tf} = \mathbf{Q}_{Cf} = 100 \mathbf{I}_{2n \times 2n}$.

\subsection{Linear Inverted Pendulum}
We consider an off-the-shelf Quanser linear inverted pendulum (cart-pole) with $n=2$ DOFs and $p=1$ actuated DOF for testing the RTC algorithm. Since, it is a commercial system, we are only interested in optimizing the swing-up controller.We additionally fix velocity related cost terms $Q_{33} = Q_{44} = 1$ to reduced the decision variables. We let a subset of hyperparameters including $Q_{11}$, $Q_{22}$ and $R_{11}$ to vary during the co-design process. As can be seen in Table \ref{tab:optSett}, the implemented algorithm resulted in an improvement of the funnel volume by a factor of 2.17. In Figure \ref{fig:funnelIncr} (left), a comparison between initial and final optimized funnels has been shown. A projection of the estimated ROA onto reduced state-spaces have been considered. Each ellipse represents the ROA associated with a knot point along the related optimal trajectory. State-space coverage improvement is evident from the overlap of the initial and final funnels. The ROA estimation has been implemented through the simulation-based method introduced in Section~\ref{sec_roaest} with a minor modification in the shrinking policy. In our implementation, just one region is shrunk at a time. The final estimated region is associated with a certificate of stability that depends on the number of simulations performed during the estimation. The verification of such a guarantee has been implemented via extensive sampling and simulation. Off-nominal initial states inside the RTC funnels are almost always stabilized. Also an experimental verification on the real system has been implemented. An impulsive torque disturbance has been introduced to check the controller's robustness. As shown in Figure \ref{fig:funnelVer} (left), the optimized set-up is able to reject the disturbance that cause the initial controller, referred to as DIRTRAN, to fail.

\subsection{Simple pendulum}
\begin{figure*}[ht]
    \begin{center}
        \includegraphics[width=0.49\textwidth]{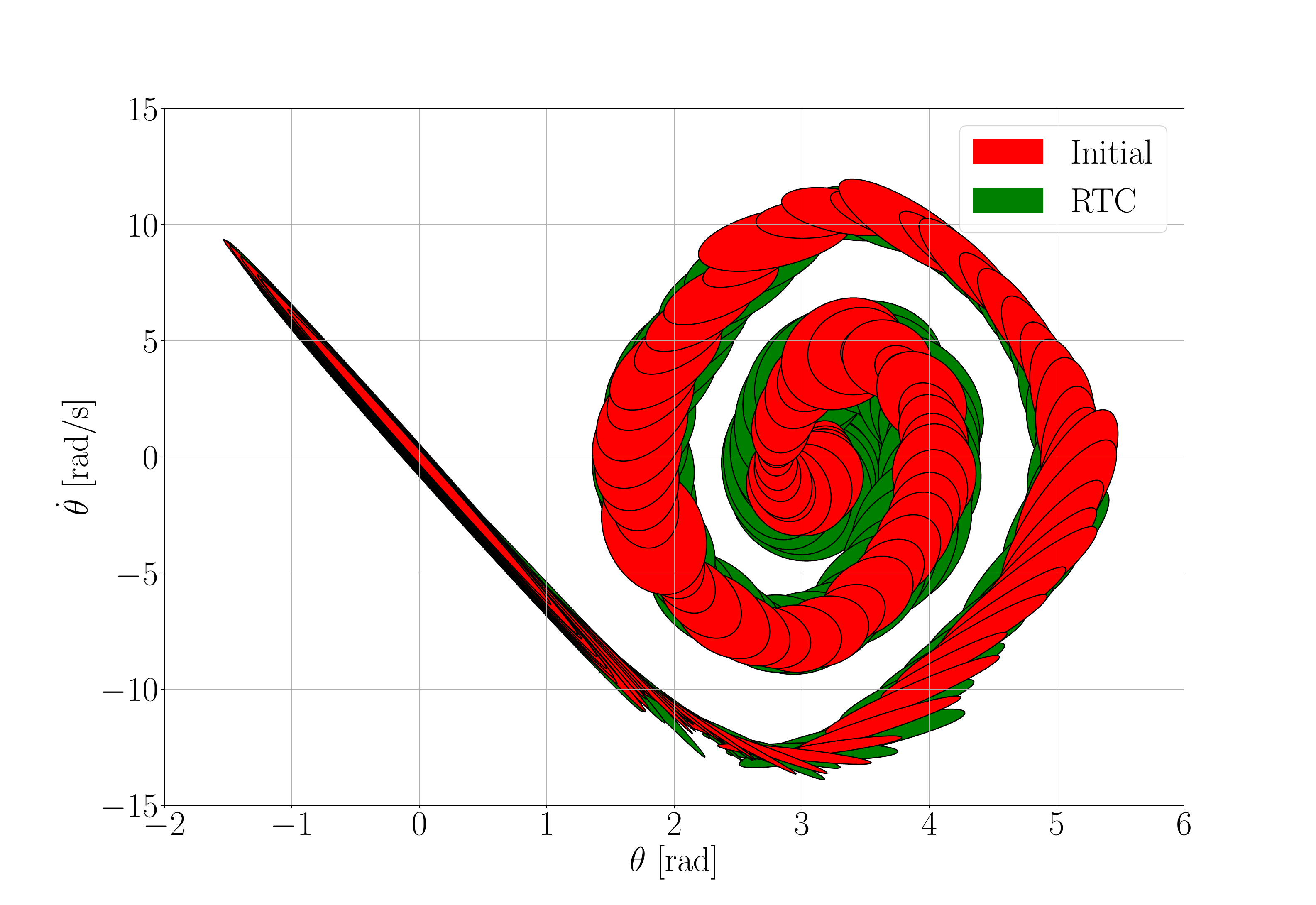}
        \includegraphics[width=0.38\textwidth]{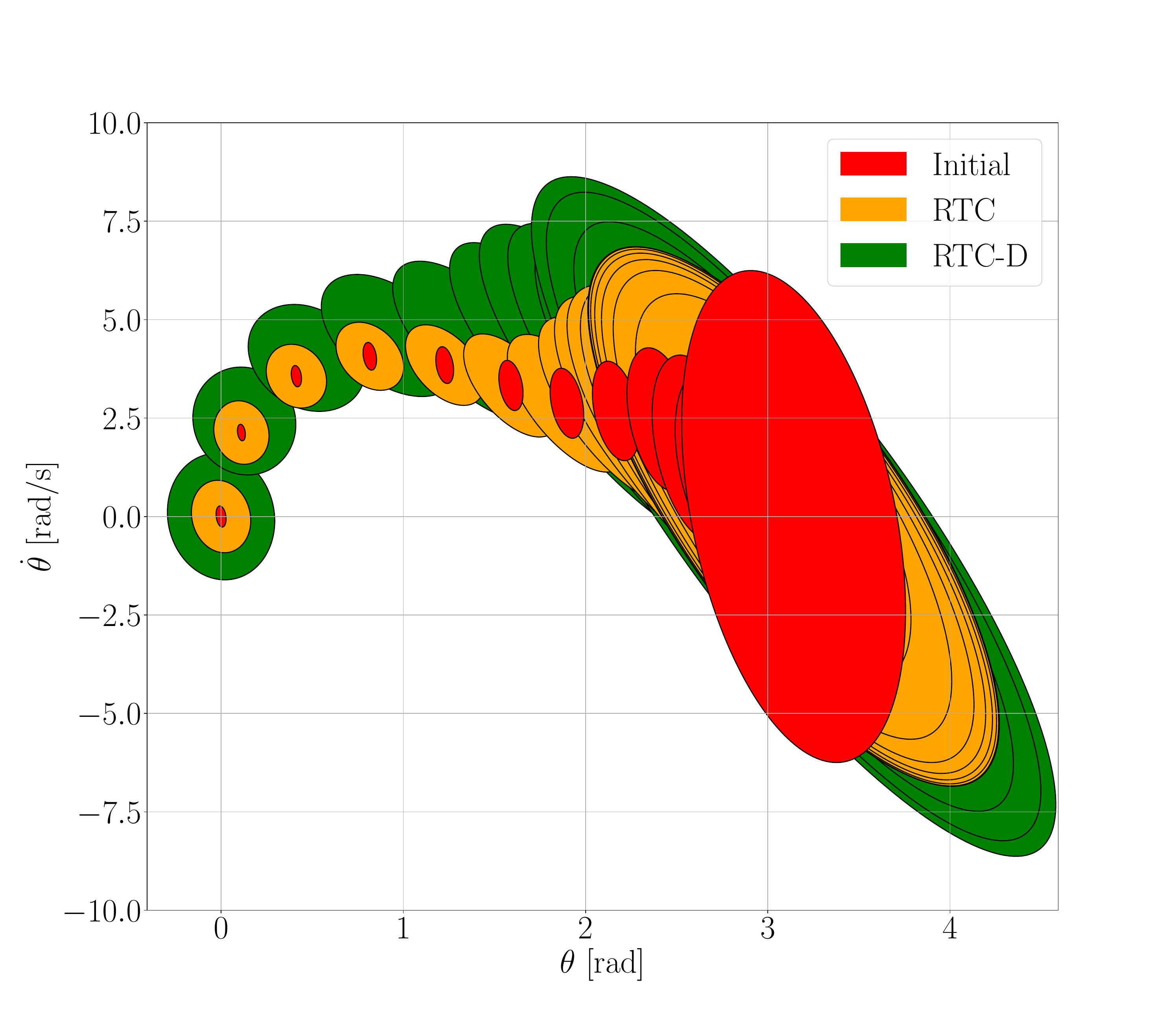}
    \end{center}
    \caption[Funnel increasing]{Funnel volume increasing due to RTC for Cart-pole (left) and comparison between RTC and RTC-D optimization  for Simple pendulum (right).}
    \label{fig:funnelIncr}
\end{figure*}

We consider a torque-limited simple pendulum with $n=p=1$ for testing the full RTC-D algorithm. The limit in the input torque renders it as one of the simplest underactuated system. Firstly, the RTC optimization is implemented with fixed design parameters, resulting in a funnel volume increase of 3.18 times as noted in Table~\ref{tab:optSett} and visually shown in Figure \ref{fig:funnelIncr} (right). Then, the mass $m$ and the link length $l$ have been chosen as decision variables for the design optimization in the outer loop. As we are aware that, in the case of this system, reducing $m$ and $l$ makes the torque limit less penalizing for the swing-up, we are expecting our algorithm to push these decision variables towards their lower limit. Applying RTC-D resulted in an optimized ROA volume that has increased to almost 4 times the initial value as reported in Table \ref{tab:optSett}. This improvement is shown in the funnel plots of Figure \ref{fig:funnelIncr} (right). RTC-D execution was stopped after almost 1500 cost function evaluations with a final execution time of 3 hours. Almost 90\% of the computational cost can be attributed to the ROA estimation phase.
As expected, the best design is the one that exhibits the least underactuation. Both values of $m$ and $l$ tend to decrease, which convince us about the sanity of the choices made by the outer loop of the codesign process.
For this system, the ROA was computed via SOS,  and the resulting region is coming with a formal guarantee of stability. Firstly, the verification of such a certificate was implemented in simulation. The simulated system's closed-loop dynamics has been tested with 1000 off-nominal initial states sampled inside the ROA. The controller was able to stabilize all of the unexpected initial states by bringing the state evolution to join the nominal trajectory.
\begin{figure*}[ht]
    \begin{center} %
        \includegraphics[width=0.44\textwidth]{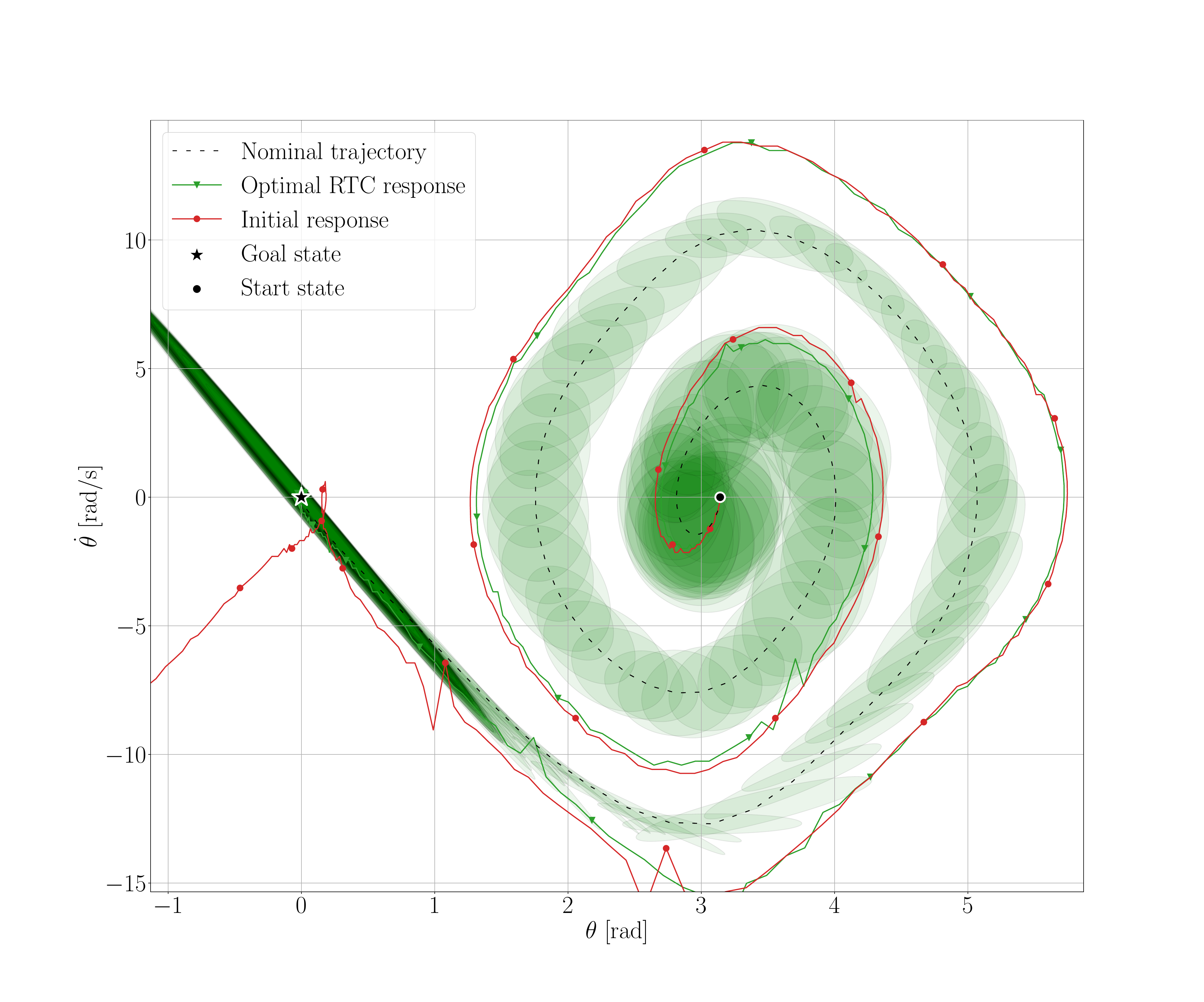}
        \includegraphics[width=0.44\textwidth]
        {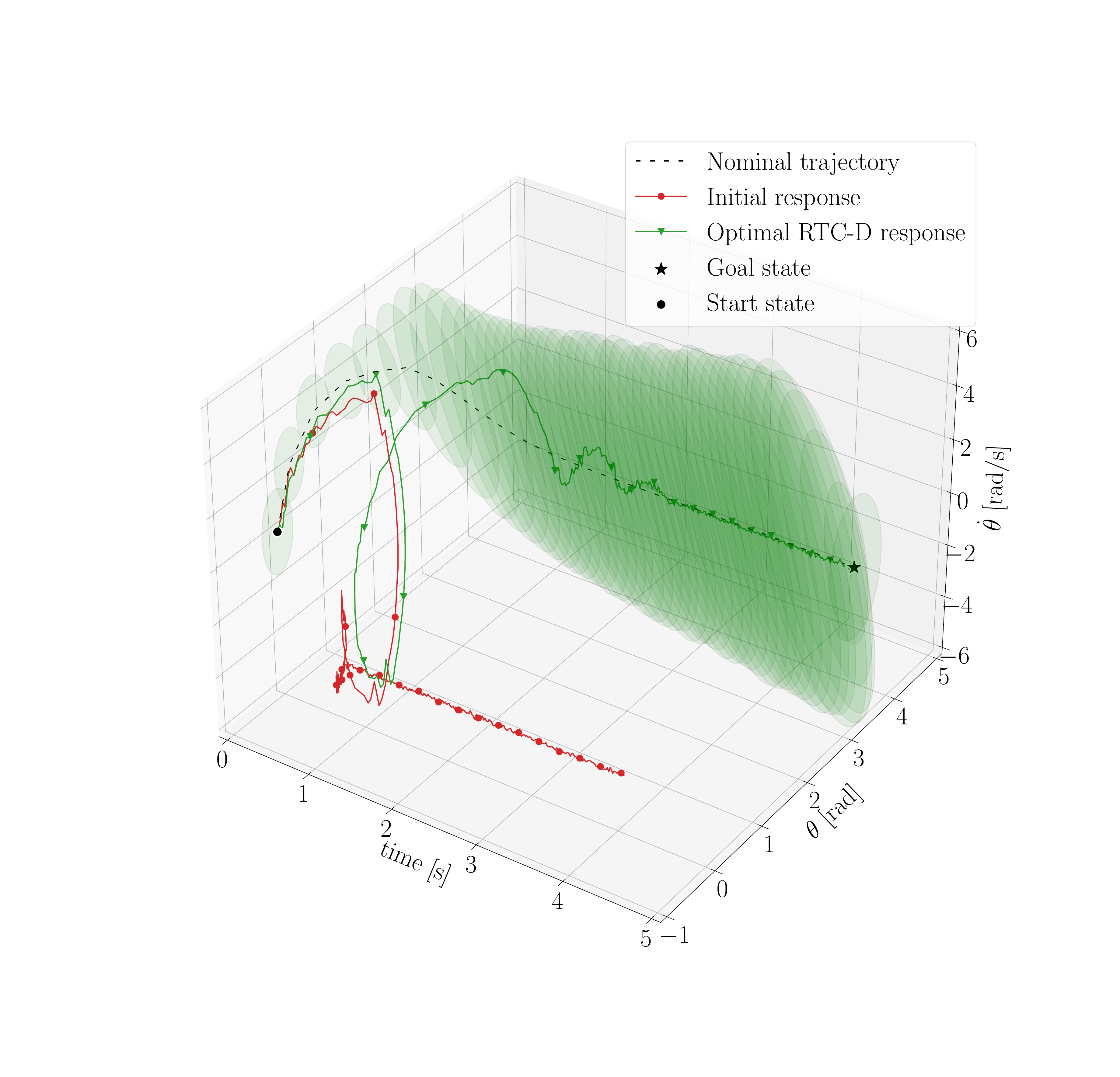}
    \end{center}
    \caption[Funnel real verification]{Experimental verification of stability guarantee (green funnel) given by the RTC for Cart-pole (left) and RTC-D for Simple pendulum (right). The optimal configuration (RTC and RTC-D) manages to achieve the desired final stabilization where the initial one does not.}
    \label{fig:funnelVer}
\end{figure*}
To assess the real world applicability of RTC-D, an experiment has also been implemented on the physical system. The initial and the optimized versions have been tested in accomplishing the swing-up task  while being affected by an impulsive torque disturbance. The experimental trajectories have been plotted with respect to the RTC-D funnel in Figure \ref{fig:funnelVer} (right). Only the initial trajectory is not able to recover the desired behaviour, i.e. to come back inside the funnel.

\section{Conclusion and Outlook}
\label{ConclusionAndOutlook}

\renewcommand{\arraystretch}{1.2}
\begin{table*}[!t]
\caption{RTC-D optimization results \label{tab:optSett}}
    \centering
    \begin{tabular}{c||ccccc|ccccc||c|c}
    \hline  
     & \multicolumn{5}{|c|}{Initial Decision Variables} & \multicolumn{5}{|c||}{Optimal Decision Variables} & \multicolumn{1}{|c|}{Volume} & \\
     System & $m\ (kg)$ & $l\ (m)$ & $Q_{11}$ & $Q_{22}$ & $R_{11}$ & $m\ (kg)$ & $l\ (m)$ & $Q_{11}$ & $Q_{22}$ & $R_{11}$ & $\ell_c^* / \ell_c^0$ & Time (h)\\ \hline  
    Cart-pole (RTC)  & 0.23  & 0.18 & 10.00  & 10.00 & 10.00  & - & -  & 10.91 & 12.57 & 6.09 & 2.17  & 3.55 \\
    Pendulum (RTC) & 0.70  & 0.40 & 10.00  & 1.00 & 0.10 & -  & - & 9.50 & 1.20 & 1.64 & 3.18 & 0.40  \\
    Pendulum (RTC-D) & 0.70  & 0.40 & 10.00  & 1.00 & 0.10 & 0.61  & 0.40 & 9.98  & 1.00 & 3.39 & 3.90 & 3.00  \\  \hline
\end{tabular}
\end{table*}

This paper presented a novel gradient free co-design algorithm, namely RTC-D, which provides optimal design parameters, nominal trajectory and stabilizing controller by taking into account their mutual dependencies. Our approach has been tested on two canonical underactuated systems: a torque-limited simple pendulum and a cart-pole. The results demonstrated an increased volume of ROA, and were validated both in simulation and through real-world experiments. 
In contrast to the approach described in \cite{FadiniRobustCoOpt}, the inner optimization layer (RTC) is implemented through DIRTRAN which allows for the introduction of explicit state and input constraints and TVLQR which allows for the incorporation of robustness analysis through time-varying ROA estimation. This approach offers an intuitive perspective and can potentially provide a stability guarantee for off-nominal states around the optimal trajectory, especially when employing the SOS-based estimation method. It's worth noting that this estimation method was not implemented in \cite{AcrobotCoDesign}, where the focus was on considering the volume of a sample-based region around the desired fixed point.  
During the empirical analysis of the optimization process, we observed that design optimization consistently mitigates underactuation. However, when dealing with the Cart-pole system, we encountered challenges due to the increased system complexity. The dimensional increase affects the computational cost of ROA estimation and, in the case of SOS-based estimation, its functionality. In our opinion, a deeper analysis on handling this problem should follow this work. As an alternative to SOS-based estimation, we used a simulation-based method. While it doesn't provide a strict formal stabilizability guarantee, it yields a viable cost for the co-optimization process. Another alternative would be to use analytical region of attraction estimation as proposed in ~\cite{2022_gross_analytical_roa}, whenever available. We then would approach the increased computational complexity with a C++ implementation and an increased use of parallelization. We also plan to extend this formulation to address the co-design of some simple hybrid dynamical systems such as AcroMonk~\cite{Javadi2023}, hopping leg~\cite{soni2023end} and RicMonk~\cite{Shourie2024}. As a future development, extending the RTC-D scheme to accommodate multiple tasks simultaneously could offer a powerful solution for more complex robotic systems such as humanoids~\cite{boukheddimi2022introducing}.

\bibliography{references}

\begin{thebibliography}{10}
\providecommand{\url}[1]{#1}
\csname url@rmstyle\endcsname
\providecommand{\newblock}{\relax}
\providecommand{\bibinfo}[2]{#2}
\providecommand\BIBentrySTDinterwordspacing{\spaceskip=0pt\relax}
\providecommand\BIBentryALTinterwordstretchfactor{4}
\providecommand\BIBentryALTinterwordspacing{\spaceskip=\fontdimen2\font plus
\BIBentryALTinterwordstretchfactor\fontdimen3\font minus
  \fontdimen4\font\relax}
\providecommand\BIBforeignlanguage[2]{{%
\expandafter\ifx\csname l@#1\endcsname\relax
\typeout{** WARNING: IEEEtran.bst: No hyphenation pattern has been}%
\typeout{** loaded for the language `#1'. Using the pattern for}%
\typeout{** the default language instead.}%
\else
\language=\csname l@#1\endcsname
\fi
#2}}

\bibitem{complexCoD}
\BIBentryALTinterwordspacing
C.~Semini, N.~G. Tsagarakis, E.~Guglielmino, M.~Focchi, F.~Cannella, and D.~G.
  Caldwell, ``Design of hyq - a hydraulically and electrically actuated
  quadruped robot,'' \emph{Proceedings of the Institution of Mechanical
  Engineers, Part I: Journal of Systems and Control Engineering}, vol. 225,
  no.~6, pp. 831--849, 2011. [Online]. Available:
  \url{https://doi.org/10.1177/0959651811402275}
\BIBentrySTDinterwordspacing

\bibitem{CoDintro}
Q.~Li, W.~Zhang, and L.~Chen, ``Design for control-a concurrent engineering
  approach for mechatronic systems design,'' \emph{IEEE/ASME Transactions on
  Mechatronics}, vol.~6, no.~2, pp. 161--169, 2001.

\bibitem{CoDex1}
J.~Allison and S.~Nazari, ``Combined plant and controller design using
  decomposition-based design optimization and the minimum principle,'' in
  \emph{Proceedings of the ASME Design Engineering Technical Conference},
  vol.~1, 01 2010.

\bibitem{CoDex2}
\BIBentryALTinterwordspacing
J.~T. Allison, T.~Guo, and Z.~Han, ``{Co-Design of an Active Suspension Using
  Simultaneous Dynamic Optimization},'' \emph{Journal of Mechanical Design},
  vol. 136, no.~8, 06 2014, 081003. [Online]. Available:
  \url{https://doi.org/10.1115/1.4027335}
\BIBentrySTDinterwordspacing

\bibitem{CoDex3}
G.~Bravo-Palacios, A.~D. Prete, and P.~M. Wensing, ``One robot for many tasks:
  Versatile co-design through stochastic programming,'' \emph{IEEE Robotics and
  Automation Letters}, vol.~5, no.~2, pp. 1680--1687, 2020.

\bibitem{CoDex4}
\BIBentryALTinterwordspacing
G.~Grandesso, G.~Bravo-Palacios, P.~Wensing, M.~Fontana, and A.~del Prete,
  ``{Exploring the limits of a hybrid actuation system through Co-Design -
  Technical Report},'' {University of Trento},'' Technical Report, June 2020.
  [Online]. Available: \url{https://hal.science/hal-02737086}
\BIBentrySTDinterwordspacing

\bibitem{limbOpt}
S.~Ha, S.~Coros, A.~Alspach, J.~Kim, and K.~Yamane, ``Task-based limb
  optimization for legged robots,'' in \emph{2016 IEEE/RSJ International
  Conference on Intelligent Robots and Systems (IROS)}, 2016, pp. 2062--2068.

\bibitem{FadiniCoOpt}
G.~Fadini, T.~Flayols, A.~Del~Prete, N.~Mansard, and P.~Souères,
  ``Computational design of energy-efficient legged robots: Optimizing for size
  and actuators,'' in \emph{2021 IEEE International Conference on Robotics and
  Automation (ICRA)}, 2021, pp. 9898--9904.

\bibitem{versatileCoD}
T.~Dinev, C.~Mastalli, V.~Ivan, S.~Tonneau, and S.~Vijayakumar, ``A versatile
  co-design approach for dynamic legged robots,'' 2022.

\bibitem{CMAES}
\BIBentryALTinterwordspacing
N.~Hansen, yoshihikoueno, ARF1, K.~Nozawa, L.~Rolshoven, M.~Chan, Y.~Akimoto,
  brieglhostis, and D.~Brockhoff, ``Cma-es/pycma: r3.2.0,'' Feb. 2022.
  [Online]. Available: \url{https://doi.org/10.5281/zenodo.6300858}
\BIBentrySTDinterwordspacing

\bibitem{salunkhe2022efficient}
D.~H. Salunkhe, G.~Michel, S.~Kumar, M.~Sanguineti, and D.~Chablat, ``An
  efficient combined local and global search strategy for optimization of
  parallel kinematic mechanisms with joint limits and collision constraints,''
  \emph{Mechanism and Machine Theory}, vol. 173, p. 104796, 2022.

\bibitem{robustMech}
A.~E. Gkikakis and R.~Featherstone, ``Robust analysis for mechanism and
  behavior co-optimization of high-performance legged robots,'' in \emph{2022
  IEEE-RAS 21st International Conference on Humanoid Robots (Humanoids)}, 2022,
  pp. 752--758.

\bibitem{DIRTREL}
\BIBentryALTinterwordspacing
Z.~Manchester and S.~Kuindersma, ``Dirtrel: Robust trajectory optimization with
  ellipsoidal disturbances and lqr feedback,'' in \emph{Robotics: Science and
  Systems (RSS)}, 2017. [Online]. Available:
  \url{https://github.com/HarvardAgileRoboticsLab/drake/tree/dirtrel}
\BIBentrySTDinterwordspacing

\bibitem{FadiniRobustCoOpt}
G.~Fadini, T.~Flayols, A.~D. Prete, and P.~Souères, ``Simulation aided
  co-design for robust robot optimization,'' \emph{IEEE Robotics and Automation
  Letters}, vol.~7, no.~4, pp. 11\,306--11\,313, 2022.

\bibitem{AcrobotCoDesign}
L.~J. Maywald, F.~Wiebe, S.~Kumar, M.~Javadi, and F.~Kirchner,
  ``Co-optimization of acrobot design and controller for increased certifiable
  stability,'' in \emph{2022 IEEE/RSJ International Conference on Intelligent
  Robots and Systems (IROS)}, 2022, pp. 2636--2641.

\bibitem{funnelIntro}
B.~R. R., R.~A. A., and K.~D. E., ``Sequential composition of dynamically
  dexterous robot behaviors,'' \emph{International Journal of Robotics
  Research}, vol.~8, no.~18, pp. 534--555, 1999.

\bibitem{MooreLQRtrees}
\BIBentryALTinterwordspacing
J.~Moore, R.~Cory, and R.~Tedrake, ``Robust post-stall perching with a simple
  fixed-wing glider using lqr-trees,'' \emph{Bioinspiration and Biomimetics},
  vol.~9, no.~2, p. 025013, may 2014. [Online]. Available:
  \url{https://dx.doi.org/10.1088/1748-3182/9/2/025013}
\BIBentrySTDinterwordspacing

\bibitem{SOSLQRtrees}
\BIBentryALTinterwordspacing
R.~Tedrake, I.~R. Manchester, M.~Tobenkin, and J.~W. Roberts, ``Lqr-trees:
  Feedback motion planning via sums-of-squares verification,'' \emph{The
  International Journal of Robotics Research}, vol.~29, no.~8, pp. 1038--1052,
  2010. [Online]. Available: \url{https://doi.org/10.1177/0278364910369189}
\BIBentrySTDinterwordspacing

\bibitem{ReistSimROA}
\BIBentryALTinterwordspacing
P.~Reist, P.~Preiswerk, and R.~Tedrake, ``Feedback-motion-planning with
  simulation-based lqr-trees,'' \emph{The International Journal of Robotics
  Research}, vol.~35, no.~11, pp. 1393--1416, 2016. [Online]. Available:
  \url{https://doi.org/10.1177/0278364916647192}
\BIBentrySTDinterwordspacing

\bibitem{2023_wiebe_doublependulum}
F.~Wiebe, S.~Kumar, L.~J. Shala, S.~Vyas, M.~Javadi, and F.~Kirchner, ``Open
  source dual-purpose acrobot and pendubot platform: Benchmarking control
  algorithms for underactuated robotics,'' \emph{IEEE Robotics and Automation
  Magazine}, pp. 2--13, 2023.

\bibitem{2018_park_tutorial}
\BIBentryALTinterwordspacing
F.~C. Park, B.~Kim, C.~Jang, and J.~Hong, ``{Geometric Algorithms for Robot
  Dynamics: A Tutorial Review},'' \emph{Applied Mechanics Reviews}, vol.~70,
  no.~1, p. 010803, 02 2018. [Online]. Available:
  \url{https://doi.org/10.1115/1.4039078}
\BIBentrySTDinterwordspacing

\bibitem{mueller2021closed}
A.~Mueller and S.~Kumar, ``Closed-form time derivatives of the equations of
  motion of rigid body systems,'' \emph{Multibody System Dynamics}, vol.~53,
  no.~3, pp. 257--273, 2021.

\bibitem{tedrakeUnderactuated}
\BIBentryALTinterwordspacing
R.~Tedrake, \emph{Underactuated Robotics}.\hskip 1em plus 0.5em minus
  0.4em\relax Course Notes for MIT 6.832, 2023. [Online]. Available:
  \url{https://underactuated.csail.mit.edu}
\BIBentrySTDinterwordspacing

\bibitem{SNOPT}
P.~E. Gill, W.~Murray, and M.~A. Saunders, ``Snopt: An sqp algorithm for
  large-scale constrained optimization,'' \emph{SIAM Review}, no. 47(1):99-131,
  2005.

\bibitem{Najafi2016}
E.~Najafi, R.~Babuška, and G.~A.~D. Lopes, ``A fast sampling method for
  estimating the domain of attraction,'' \emph{Nonlinear Dynamics}, no.~86, p.
  823–834, 2016.

\bibitem{KhalisBook}
H.~K. Khalil, \emph{Non linear Systems}.\hskip 1em plus 0.5em minus 0.4em\relax
  Prentice Hall, Upper Saddle River, NJ, 1996.

\bibitem{Moore_2014}
\BIBentryALTinterwordspacing
J.~Moore, R.~Cory, and R.~Tedrake, ``Robust post-stall perching with a simple
  fixed-wing glider using lqr-trees,'' \emph{Bioinspiration \& Biomimetics},
  vol.~9, no.~2, p. 025013, may 2014. [Online]. Available:
  \url{https://dx.doi.org/10.1088/1748-3182/9/2/025013}
\BIBentrySTDinterwordspacing

\bibitem{Sproc}
S.~Boyd and L.~Vandenberghe, ``Convex optimization,'' cambridge University
  Press, p. 655.

\bibitem{DFKIsp}
\BIBentryALTinterwordspacing
F.~Wiebe, J.~Babel, S.~Kumar, S.~Vyas, D.~Harnack, M.~Boukheddimi, M.~Popescu,
  and F.~Kirchner, ``Torque-limited simple pendulum: A toolkit for getting
  familiar with control algorithms in underactuated robotics,'' \emph{Journal
  of Open Source Software}, vol.~7, no.~74, p. 3884, 2022. [Online]. Available:
  \url{https://doi.org/10.21105/joss.03884}
\BIBentrySTDinterwordspacing

\bibitem{QuanserLIP}
\BIBentryALTinterwordspacing
Quanser, ``Linear servo base unit with inverted pendulum,'' simulink
  Courseware. [Online]. Available:
  \url{https://www.quanser.com/products/linear-servo-base-unit-inverted-pendulum/}
\BIBentrySTDinterwordspacing

\bibitem{2022_gross_analytical_roa}
L.~Gross, L.~Maywald, S.~Kumar, F.~Kirchner, and C.~Lüth, ``Analytic
  estimation of region of attraction of an lqr controller for torque limited
  simple pendulum,'' in \emph{2022 IEEE 61st Conference on Decision and Control
  (CDC)}, 2022, pp. 2695--2701.

\bibitem{Javadi2023}
\BIBentryALTinterwordspacing
M.~Javadi, D.~Harnack, P.~Stocco, S.~Kumar, S.~Vyas, D.~Pizzutilo, and
  F.~Kirchner, ``{AcroMonk}: A minimalist underactuated brachiating robot,''
  \emph{{IEEE} Robotics and Automation Letters}, vol.~8, no.~6, pp. 3637--3644,
  jun 2023. [Online]. Available:
  \url{https://doi.org/10.1109\%2Flra.2023.3269296}
\BIBentrySTDinterwordspacing

\bibitem{soni2023end}
R.~Soni, D.~Harnack, H.~Isermann, S.~Fushimi, S.~Kumar, and F.~Kirchner,
  ``End-to-end reinforcement learning for torque based variable height
  hopping,'' in \emph{2023 IEEE/RSJ International Conference on Intelligent
  Robots and Systems (IROS)}.\hskip 1em plus 0.5em minus 0.4em\relax IEEE,
  2023, pp. 7531--7538.

\bibitem{Shourie2024}
G.~S. Shourie, M.~Javadi, S.~Kumar, H.~Z. Boroujeni, and F.~Kirchner,
  ``Ricmonk: A three-link brachiation robot with passive grippers for
  energy-efficient brachiation,'' in \emph{2024 International Conference on
  Robotics and Automation (ICRA)}, 2024.

\bibitem{boukheddimi2022introducing}
M.~Boukheddimi, S.~Kumar, H.~Peters, D.~Mronga, R.~Budhiraja, and F.~Kirchner,
  ``Introducing rh5 manus: A powerful humanoid upper body design for dynamic
  movements,'' in \emph{2022 International Conference on Robotics and
  Automation (ICRA)}.\hskip 1em plus 0.5em minus 0.4em\relax IEEE, 2022, pp.
  01--07.

\end{thebibliography}

\end{document}